\def\year{2021}\relax
\documentclass[letterpaper]{article} %
\usepackage{aaai21}  %
\usepackage{times}  %
\usepackage{helvet} %
\usepackage{courier}  %
\usepackage[hyphens]{url}  %
\usepackage{graphicx} %
\usepackage{natbib}  %
\usepackage{caption} %
\newcommand{\ours}{\textit{SCP}}
\newcommand{\ourtitle}{Learning Complex 3D Human Self-Contact}

\usepackage{graphicx}
\usepackage{multirow}
\usepackage{comment}
\usepackage{amsmath,amssymb} %
\usepackage{color}
\usepackage{makecell}
\usepackage[switch]{lineno}  
\usepackage{url}

\title{\ourtitle}

\author {
    Mihai Fieraru\textsuperscript{\rm 1} \quad 
    Mihai Zanfir\textsuperscript{\rm 1} \quad 
    Elisabeta Oneata\textsuperscript{\rm 1} \quad \\
    Alin-Ionut Popa\textsuperscript{\rm 1} \quad 
    Vlad Olaru\textsuperscript{\rm 1} \quad 
    Cristian Sminchisescu\textsuperscript{\rm 2,1} \\
}
\affiliations {
    \textsuperscript{\rm 1} Institute of Mathematics of the Romanian Academy,
    \textsuperscript{\rm 2} Lund University\\
    {\tt\small \textsuperscript{\rm 1}\{firstname.lastname\}@imar.ro, \textsuperscript{\rm 2}cristian.sminchisescu@math.lth.se}
}

\begin{document}

\maketitle

\begin{abstract}
Monocular estimation of three dimensional human self-contact is fundamental for detailed scene analysis including body language understanding and behaviour modeling. 
Existing 3d reconstruction methods do not focus on body regions in self-contact and consequently recover configurations that are either far from each other or self-intersecting, when they should just touch. This leads to perceptually incorrect estimates and limits impact in those very fine-grained analysis domains where detailed 3d models are expected to play an important role. To address such challenges we detect self-contact and design 3d losses to explicitly enforce it. Specifically, we develop a model for \underline{S}elf-\underline{C}ontact \underline{P}rediction (\ours{}), that estimates the body surface signature of self-contact, leveraging the localization of self-contact in the image, during both training and inference. We collect two large datasets to support learning and evaluation: (1) HumanSC3D, an accurate 3d motion capture repository containing $1,032$ sequences with $5,058$ contact events and $1,246,487$ ground truth 3d poses synchronized with images collected from multiple views, and (2) FlickrSC3D, a repository of $3,969$ images, containing $25,297$ surface-to-surface correspondences with annotated image spatial support. We also illustrate how more expressive 3d reconstructions can be recovered under self-contact signature constraints and present monocular detection of face-touch as one of the multiple applications made possible by more accurate self-contact models. 
\end{abstract}

\section{Introduction}

Most monocular 3d human reconstruction systems do not directly infer human self-contact, although its central role in correctly recognizing the subtleties of many iconic poses or gestures is widely acknowledged perceptually. 
Current modeling deficiencies result in contact regions either far from each other or self-intersecting in the final 3d reconstruction, when they should instead just touch (e.g. contact between one's hand and chin). In turn, unpredictable reconstructions of self-contact decrease the appeal of using 3d representations for fine grained analysis of behavior and intent, particularly as many self-touch events are elicited frequently, and with little or no human awareness. Correctly tracking self-contact would be invaluable not just for behavior analysis but in assessing hygiene and possible health implications during a pandemic. 
\begin{figure}[!t]
\begin{center}
        \includegraphics[width=.8\linewidth]{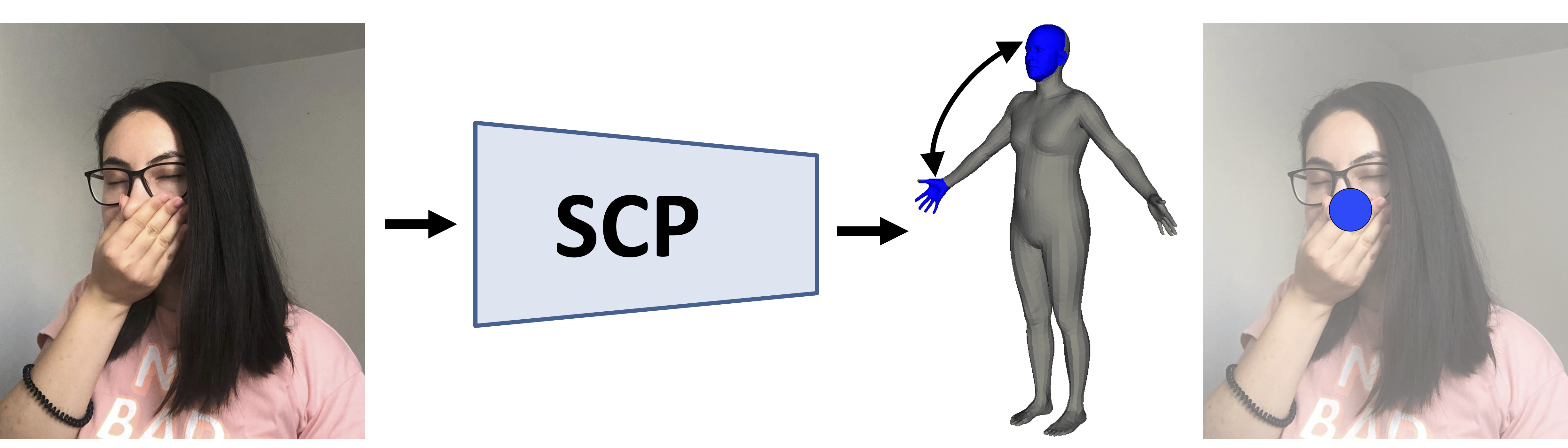}
\end{center}
\caption{Our self-contact prediction network (\ours{}) estimates the body regions in contact, their correspondences and the self-contact positioning in image space.}
\end{figure}

To overcome some of the shortcomings of existing, self-contact agnostic, 3d reconstruction methods, we propose to represent self-contact explicitly and show how the resulting models can assist behavioural understanding in applications assessing face touching. Our models learn to predict the image location of contact in order to assist the detection of body regions in self-contact, as well as their signature, defined as the correspondences between regions on the surface of a human body model that touch. Conditioned on such detailed estimates self-contact can be recovered correctly in the 3d reconstruction. To train models and for large-scale quantitative evaluation, we collect and annotate two large scale datasets containing images of people in self-contact. HumanSC3D is an accurate 3d motion capture dataset containing $1,032$ sequences with $5,058$ contact events and $1,246,487$ ground truth 3d poses synchronized with images captured from multiple views. We also collect FlickrSC3D, a dataset of $3,969$ images, containing $25,297$ annotations of body part region pairs in contact, defined on a 3d human surface model, together with their self-contact localisation in the image. 

The main contributions of the paper are as follows:
\begin{itemize}
    \item Introduce a first principled model to detect self-contact body regions and their signature. Our novel deep neural network \ours{} is assisted by an intermediate self-contact image localisation (branch) predictor, leveraged both in training, for local feature selection, and in testing, by enforcing consistency with the estimated 3d contact signature.
    \item Novel, task-specific, large scale, valuable community datasets capturing people in self-contact, together with dense annotations of a 3d body model to capture the surface regions in contact, as well as image annotations associated to the observed points of contact. The data and models will be made available for research.
    \item Quantitative and qualitative demonstration of metrically more accurate and perceptually veridical 3d reconstructions based on self-contact signatures.
    \item A foundation for a large class of applications that would benefit from accurate 3d self-contact representations, such as, health monitoring of possible infections when hands touch parts of the face (mouth, nose, eyes) in hospitals or during a pandemic, or subtle behavioral understanding of gestures for robot-assisted therapy of children with autism, to name just a few.
\end{itemize}
 
\section{Related Work}

Automatic 3d human pose and shape estimation from images and video has been increasingly studied in recent years and significant progress has been made 
\cite{Mehta_3DV2018,zanfir_nips2018,li2019crowdpose,su2019multi,benzine2019deep,KolotourosIccv2019,humanMotionKanazawa19,kocabas2019vibe}. These methods focus on 3d pose, and to some extent shape estimation, and person's relative placement with respect to the scene. However, the subtleties of 3d shape especially in conjunction with contact are still largely unexplored, with vast potential for improvement well beyond existing art. Challenges include human-object interaction, inter-human interactions or human self-contact. In this paper we present models and insights - methodological, experimental and logistic, in terms of data collection - focusing on human self-contact. In the rest of this section we review previous work on human contact and self-contact applications.   

\noindent{\bf Self-Contact.} Most of previous work on self-contact \cite{Tzionas2016,GCPR_2013_Tzionas_Gall,Taylor2017,mueller_siggraph2019} applies to the interaction of human extremities, such as hands. \citet{Tzionas2016} introduces a method for modelling 3d hand to hand or hand to object interactions based on RGB-D data. The hand reconstruction is done via an energy-based modeling which incorporates physics and collision constraints. However the shape of the hand is not estimated, and only a standard template is used. \citet{mueller_siggraph2019} propose a similar real-time system based on a RGB-D sensor that is also able to estimate the shape and pose of interacting hands. None of the above methods explicitly detects the regions in self-contact or predicts their signature. In contrast, we handle full bodies, not only hand regions, and do not require depth data.
Others \cite{bogo2016, Zanfir_2018_CVPR, pavlakos2019expressive} use non-self intersection constraints to prevent inadmissible 3d human reconstructions. Avoiding self-collisions only, and in the absence of any semantics of self-contact and the underlying surface regions, however, makes it difficult to enforce self-contact of surfaces when these actually touch. 

\noindent{\bf Human - Object/Scene Contact.} Contact between humans and the environment has also been studied recently. \citet{Hasssan2019Prox} propose an optimisation method for 3d human shape estimation which incorporates scene constraints (including a contact-aware loss function) in the form of depth information. Leveraging the same contact loss, \cite{zhang2020generating} learn how to plausibly place 3d people in 3d scenes. Contact between human feet and the ground is also modeled in \cite{zou2020reducing}, as it has been previously done by \cite{Zanfir_2018_CVPR}, and used to constrain 3d human reconstruction.
Interaction with objects is also studied in \cite{Hasson_2019_CVPR}, who jointly model the reconstruction of human hands and interacting objects based on single view RGB data. \citet{li2019estimating} estimates contact positions, forces and torques.

\noindent{\bf Human - Human Contact.} Contact between people is typical in close interactions like business meetings, informal conversations, or other social events. \citet{liu2013markerless,liu2011markerless} scan participants and rig them to a 3d skeleton. Given a green background setup, the motion in various scenarios is recorded and later estimated based on an energy model. Yet, the interaction between participants is modelled solely by non-self intersection constraints. In our recent work \cite{fieraru2020cvpr} we model and learn contact regions between  people by means of their contact signature. However, we did not cover self-contact and the image localization of contact is not annotated or estimated. Still, in this work we also consider as baseline \cite{fieraru2020cvpr}'s ISP prediction method, adapted to estimate self-contact.

\noindent{\bf Applications.} Self-contact prediction can enable numerous applications, and we only reference a few here. \citet{kwok2015face} study hygiene and virus transmission, by monitoring how often students touch their face with their hands. Yet, data is gathered manually with multiple investigators annotating videotape recordings. \citet{mueller2019self} also study the locations and durations of facial self-touches, but using accelerometers and EMG. An automatic labeling approach such as our \ours{} has the potential to enable the automation of larger such quantitative studies using only RGB sensors. 
Gesture analysis can benefit from improved self-contact signature predictions. For example, applause detection \cite{applausedetection} can possibly be performed from soundless videos as the detection of frequent self-contact between one's hands. Similarly, a self-contact signature such as covering ears with both hands can be used as a signal of patients' noise-sensitivity, with applications in robot-assisted autism therapy \cite{rudovic2017measuring,marinoiu18deenigma}.

\section{Methodology}
\begin{figure*}[!htbp]
\begin{center}
         \includegraphics[width=0.87\linewidth]{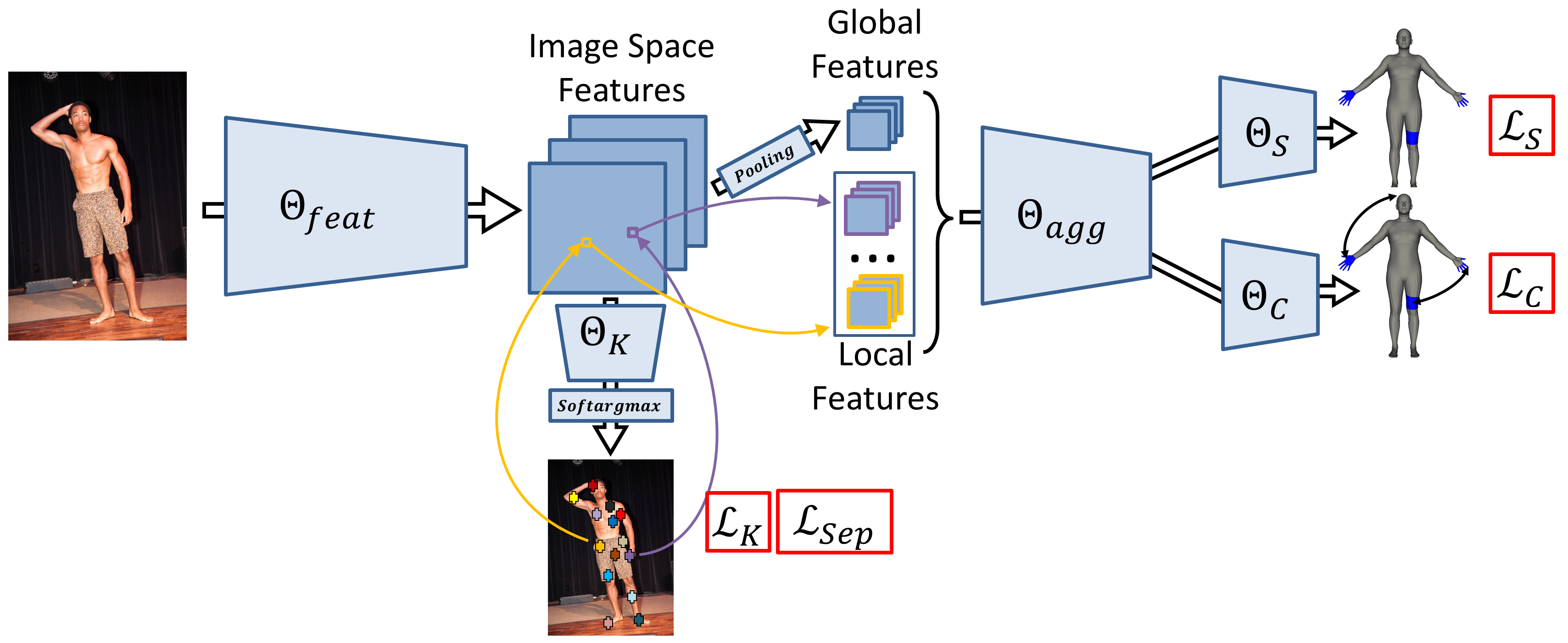}
\end{center}
\caption{Our \ours{} architecture that estimates self-contact spatial support $K$, supervised by both ${\mathcal{L}}_K$ (eq.~\ref{eq:kloss}) and ${\mathcal{L}}_{sep}$ (eq.~\ref{eq:seploss}), and self-contact segmentation $S$ and signature $C$, supervised by losses ${\mathcal{L}}_S$ and ${\mathcal{L}}_C$. The input is an RGB image cropped around a person. \ours{} predicts the spatial support of self-contact $K$ with $\Theta_{K}$ and uses it to select local features (one for each body region). Merged with global features, these are processed by an aggregation layer $\Theta_{agg}$ and specialization layers, for segmentation  $\Theta_S$ and signature prediction $\Theta_C$.}
\label{fig:pipeline_overview}
\end{figure*}

In this section, we describe our proposed model \ours{} to predict the image spatial support of self-contact, the self-contact segmentation and the self-contact signature, as well as our model for 3d human reconstruction under self-contact constraints.
Building up on our earlier work on modeling contact between people \cite{fieraru2020cvpr}, we define the self-contact segmentation and signature of a person in an image $I$ by 
discretizing of the surface of the human body model into $N_{R}$ regions. In our case, region-level self-contact \textbf{signature} $C^{R}(I) \in \{0, 1\}^{N_{R}} \times \{0, 1\}^{N_{R}}$ is defined as $C^{R}_{r_1, r_2}(I) = 1$ when region $r_1$ is in contact with region $r_2$, where $r_1 \neq r_2$ are surface regions of the same person and $C^{R}_{r_1, r_2}(I) = 0$ otherwise. Note that $C^{R}(I)$ is a symmetric matrix. Similarly, region-level self-contact \textbf{segmentation} $S^{R}(I) \in \{0, 1\}^{N_{R}}$ is defined as $S^{R}_{r}(I) = 1$ when $r$ is in contact with any other surface region on the same body and $S^{R}_{r}(I) = 0$ otherwise. 
In addition, we introduce the notion of \textbf{image support} of a region self-contact:
\begin{equation}
K^{R}(I) = \{(x_r, y_r) | S^{R}_{r}(I)=1\}
\end{equation}
where $[x_r, y_r]$ is the coordinate of the center of region $r$ projected in the image.

An overview of \ours{} is illustrated in fig.~\ref{fig:pipeline_overview}.
\ours{} takes as input an RGB image cropped around a person and learns to extract image space features  $\Theta_{feat}$ using the backbone of the ResNet50 \cite{he2016deep} architecture, up to the 16th convolutional layer. 

\subsection{Self-Contact Image Support}
One way in which the image support of self-contact can be leveraged is by informing the selection of local features needed for downstream tasks. To this end, after the feature encoder $\Theta_{feat}$, we extract a set of $N_{R}$ heatmaps using $\Theta_K$ and apply the softargmax operation to obtain a set of $N_{R}$ image coordinates $\{(\widetilde{x_r}, \widetilde{y_r})| r \in \{1,\ldots, N_{R}\}\}$. 

For regions that belong to the image support of self-contact ground truth $K^{R}(I)$, we apply the Euclidean loss $\mathcal{L}_K$ to guide the discovered landmarks towards the ground-truth image support. Note that corresponding surface regions in self-contact have the same spatial support.
\begin{equation}\label{eq:kloss}
    \mathcal{L}_K = \frac{1}{|K^{R}(I)|}\sum_{\substack{(x_r, y_r) \\ \in K^{R}(I)}} \| (x_r, y_r) - (\widetilde{x_r}, \widetilde{y_r})\|_2^2
\end{equation}

For pairs of regions not in self-contact, we impose a separation constraint $\mathcal{L}_{sep}$ to guide them towards different image areas. We adopt the loss function proposed in \cite{zhang2018unsupervised}, that highly penalizes small distances between points and vanishes quickly as distance between them increases.
\begin{equation}\label{eq:seploss}
    \mathcal{L}_{sep} = \sum_{\substack{(r_1, r_2)\\ C^{R}_{r_1, r_2}(I)=0}} \exp \bigg( \frac{\| (\widetilde{x_{r_1}}, \widetilde{y_{r_1}}) - (\widetilde{x_{r_2}}, \widetilde{y_{r_2}})\|_2^2}{-2\sigma_{sep}^2 }\bigg)
\end{equation}
One can observe that $\mathcal{L}_{sep}$ is a weakly-supervised loss, since it does not require ground truth support $K^{R}(I)$, but only ground truth self-contact signatures $C^{R}(I)$.

\subsection{Self-Contact Segmentation and Signature}
The network can use the $N_{R}$ landmarks as putative locations where local image features can be extracted. For each region $r$, we sample the image space features at location  $(\widetilde{x_r}, \widetilde{y_r})$. Since $(\widetilde{x_r}, \widetilde{y_r})$ are continuous, we bilinearly interpolate between features at the $4$ nearby discrete coordinates on the image space grid. Each of the $N_{R}$ local features are then concatenated with global features (obtained by a holistic pooling operation on the image space features) and fed to an aggregation module $\Theta_{agg}$. This is implemented as a series of two fully connected layers that progressively reduce the dimensionality of the $N_{R}$ features.

For self-contact segmentation and signature tasks, we draw inspiration from the two specialization layers and the underlying losses introduced in \cite{fieraru2020cvpr}. $\Theta_{S}$ and $\Theta_{C}$ are fully connected layers, and $\mathcal{L}_{S}, \mathcal{L}_{C}$ are sigmoid cross-entropy losses, with the positive and negative classes appropriately weighted. While $\mathcal{L}_{S}$ is applied
directly on the output of $\Theta_{S}$, for the case of correspondences, the loss $\mathcal{L}_{C}$ is applied on the feature similarity matrix $FF^T$, where $F$ is the output of $\Theta_{C}$. We also experiment with the Euclidean distance as an alternative similarity metric to the dot product used in $FF^T$, but confirm experimentally that it does not outperform it. 

At inference, we propose, as novelty, to use both the estimated self-contact segmentation and the self-contact image support in order to limit erroneous correspondences in the predicted self-contact signature, as this has a large output space and is difficult to learn. First, we remove all correspondences involving regions not found in the predicted segmentation. Second, we remove all predicted correspondences between two regions whose estimated landmarks are not close to each other. This enforces consistency of signature with the image support, since two regions in correspondence should also have their image projection in proximity.

\subsection{Self-Contact Signatures for 3D Reconstruction}
Self-contact signatures are also used to constrain 3d human reconstruction to be consistent. We showcase this using the optimization framework of \cite{Zanfir_2018_CVPR} augmented in \cite{fieraru2020cvpr} with interaction contact signature losses. The cost function, adapted for the reconstruction of a single person, and by using self-contact consistency, becomes:
\begin{equation}
\label{eq:full_loss}
  L = L_{S} + L_{psr} + L_{col} + L_{G}
\end{equation}
where $L_{S}$ is the projection error with respect to estimated semantic body part labeling and 2d body pose, $L_{psr}$ is a pose and shape regularization cost, and $L_{col}$ is a self-collision penalty term. $L_{G} = L_{D} + L_{N}$ is adapted to be a contact consistency cost for self-contact signature, where $L_{D}$ minimizes the distance between pairs of regions in self-contact and $L_{N}$ is a term aligning the orientation of region surfaces found in self-contact. 

Please check our Sup. Mat. for further details on both the \ours{} network and the optimization framework.
\section{Proposed Datasets}
\noindent{\bf HumanSC3D.} As current 3d human pose datasets such as Human3.6m \cite{Ionescu14pami} or 3DPW \cite{vonMarcard2018} contain relatively few frames in self-contact, to evaluate our proposed methodology, we collect a new dataset of people in more challenging self-contact poses. 

HumanSC3D contains 3d motion capture data of $6$ human participants ($3$ men and $3$ women between 20 and 30 years old with various fitness levels and body shapes), and videos captured by $4$ synchronized RGB cameras. The subjects are shown a series of images with ordinary poses and self-contact and asked to reproduce only the type of contact they see (not the pose), such as: touching one's chin, crossing one's legs or arms, etc. In addition, they are instructed to continuously change their body position and orientation relative to the cameras for increased variability. For each scenario, we record a short clip that captures the transition from an A-pose to the desired self-contact and back to the A-pose. In total, each subject performs $172$ motions, out of which $116$ are standing, $20$ sitting on the floor and $36$ sitting on or standing next to a chair, summing up to a total of $1,246,487$ ground truth 3d poses and associated RGB images.

We also capture each subject's body shape using a 3d body scanner. By fitting our body model to the body scans for shape and to the 3d marker positions for pose, we also obtain (pseudo) ground truth reconstructions \cite{xu2020ghum}.

For each of the $172$ self-contact scenarios of any given subject we extract a middle frame (where the person is in self-contact) and manually label the body regions in contact and their correspondence. The annotation is performed by clicking on the surface of a 3d human body model with $\sim10k$ vertices. In addition, the annotators are also asked to roughly indicate the spatial support of the self-contact in the image, by clicking in the original image at the position of each self-contact. In this way, we obtain $4,128$ images with people and associated self-contact information (multiple correspondences between two facets of the mesh and pixels of the image). Although in a controlled environment, we choose to manually annotate self-contact for higher quality control. Alternative approaches, such as multi-view marker-based reconstruction, can still fail under complex self-contact, especially for the hand regions where markers are sparsely placed or occluded. 

\begin{figure}[!htbp]
\begin{center}
\def\th{46pt}
\def\bh{67pt}
\scalebox{0.97}{
\begin{tabular}{ccc}
        \textbf{contact} & \textbf{no contact} & \textbf{uncertain contact} \\
         \includegraphics[height=\th]{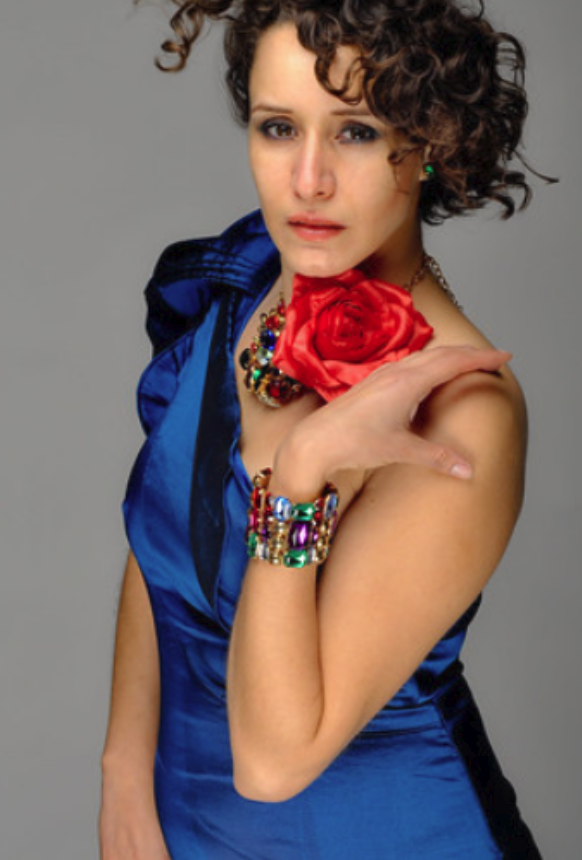}
         \includegraphics[height=\th]{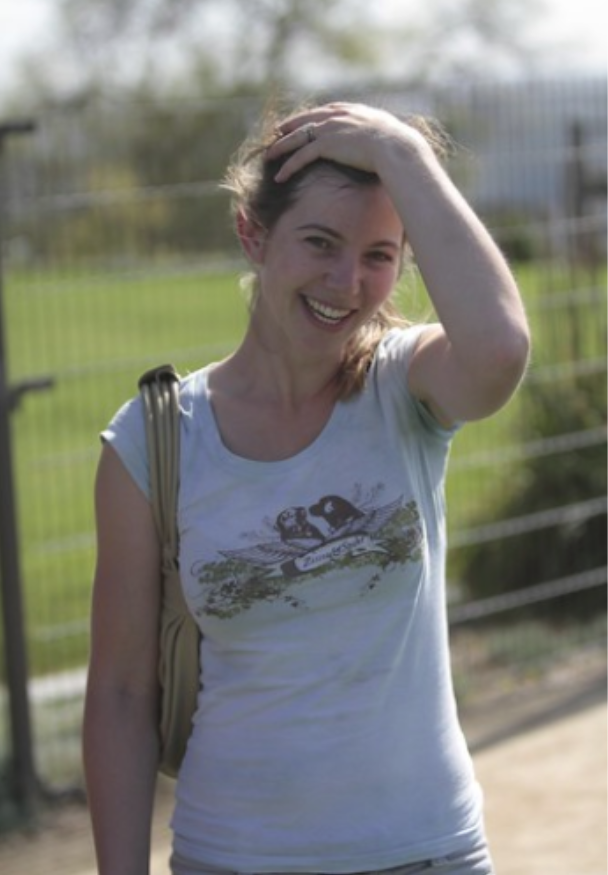} &
         \includegraphics[height=\th]{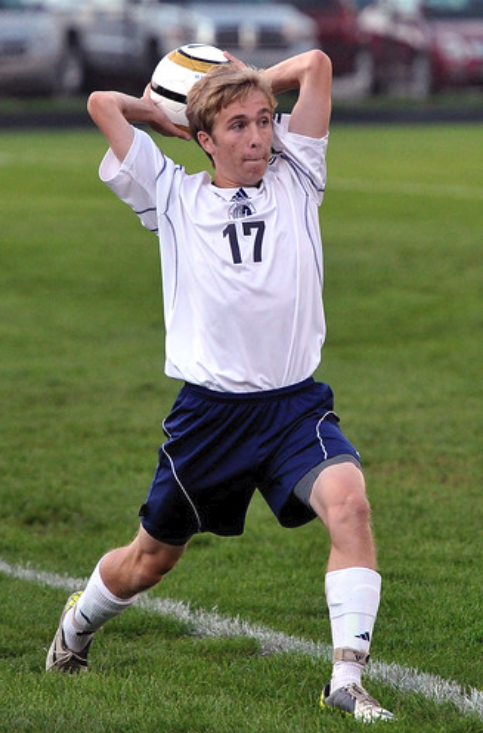}
         \includegraphics[height=\th]{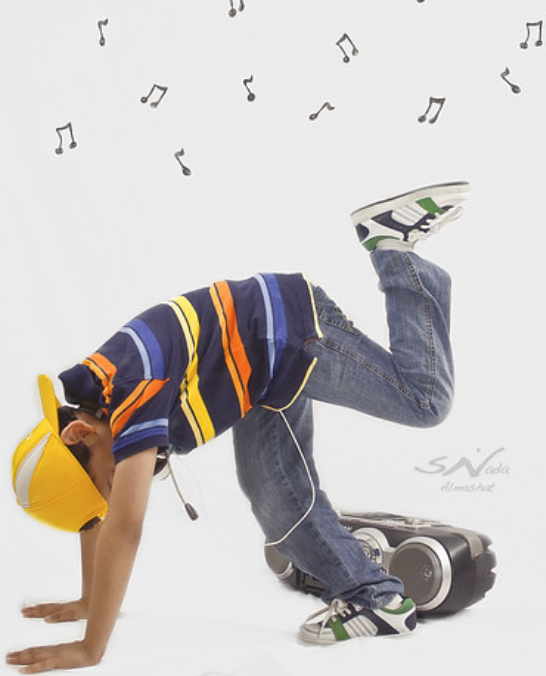} &
         \includegraphics[height=\th]{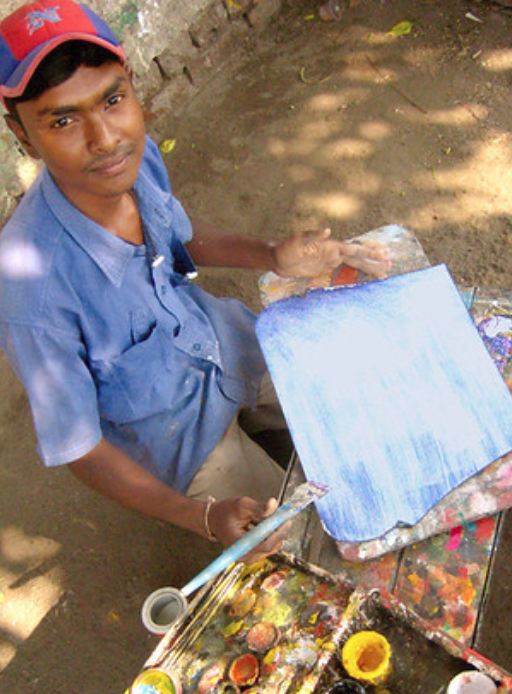}
         \includegraphics[height=\th]{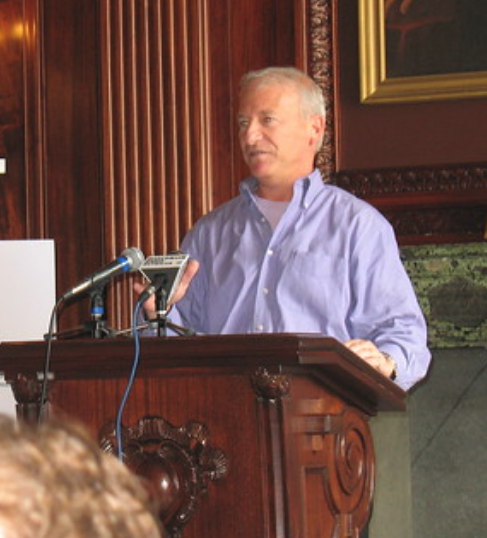} \\
        \includegraphics[height=\bh]{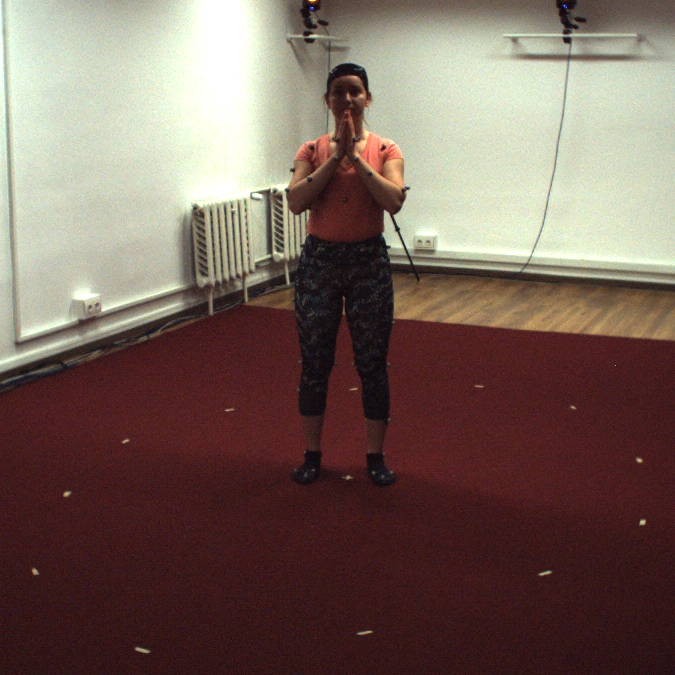} &
         \includegraphics[height=\bh]{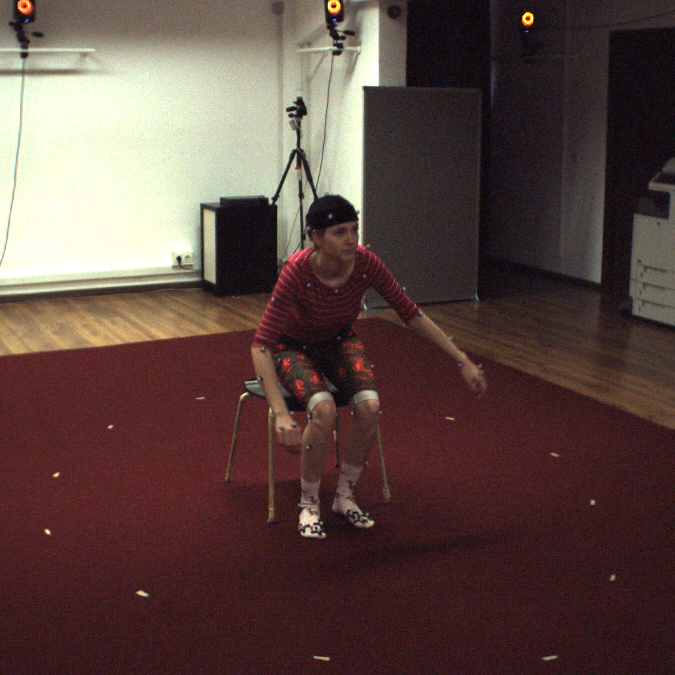} &
        \includegraphics[height=\bh]{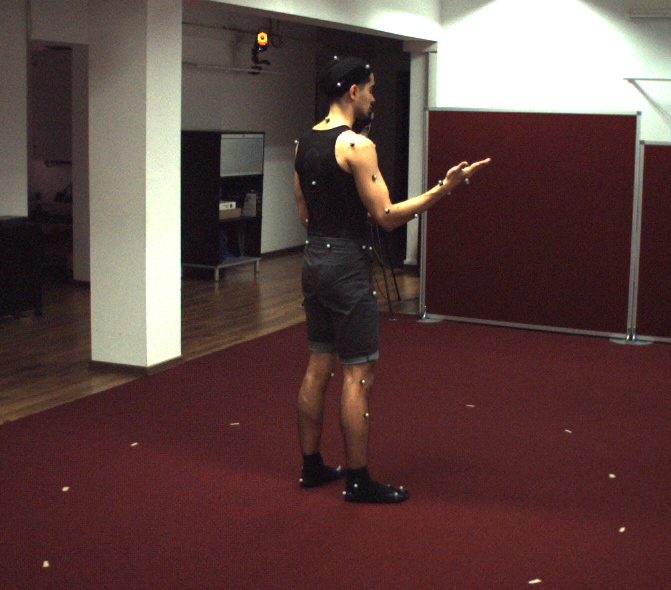}      
\end{tabular}}
\end{center}
\caption{Examples of images from our FlickrSC3D \textbf{(top)} and HumanSC3D \textbf{(bottom)} dataset. \textbf{Left}: People in self-contact. \textbf{Center}: People not in self-contact. \textbf{Right}: Uncertain whether the person is in self-contact or not.}
\label{fig:flickr_example}
\end{figure}

\begin{figure}[!htbp]
\begin{center}
\includegraphics[height=0.15\textheight]{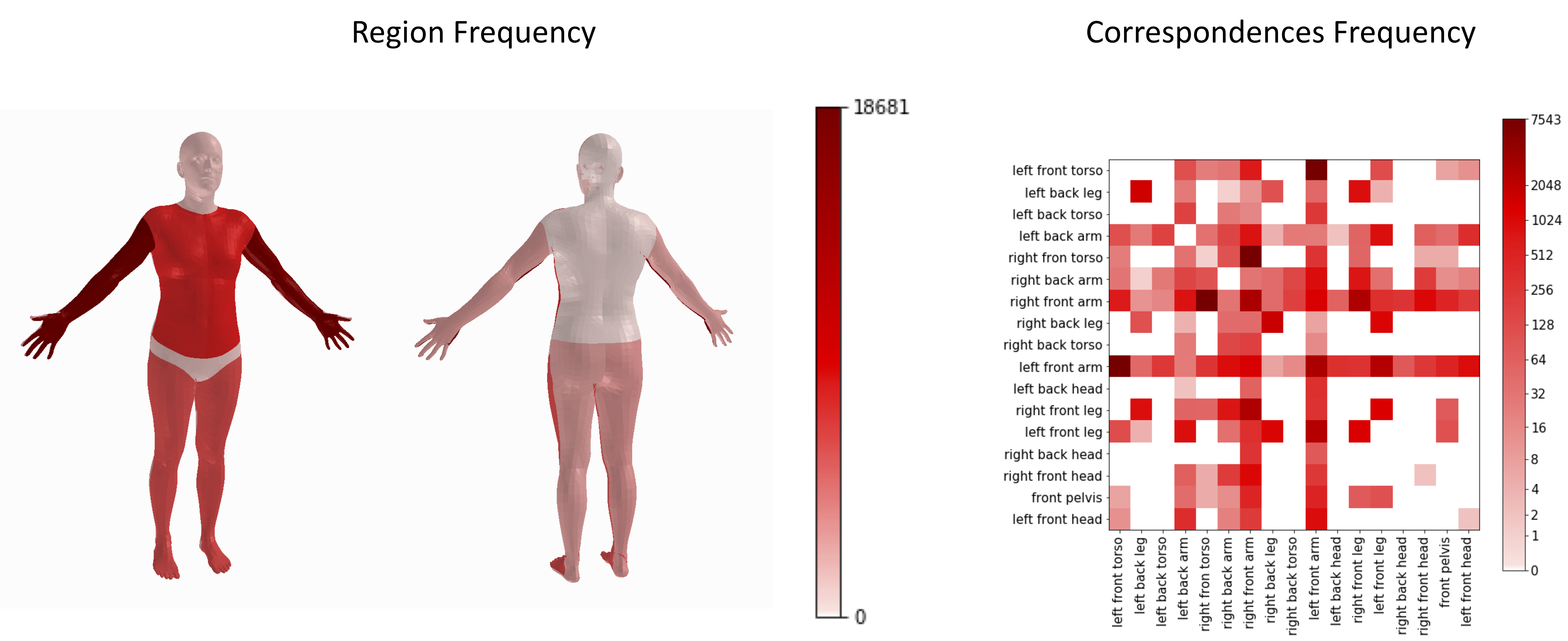} 
\end{center}
\caption{\textbf{Left}: Body region frequency of self-contact (75 regions). Note the left-right symmetry and the high frequency for the arms, hands, legs and torso regions. \textbf{Right}: Self-contact correspondence counts (17 regions).}
\label{fig:contact_statistics}
\end{figure}

\noindent{\bf FlickrSC3D.} To further extend our experiments to natural settings, we gather $3,969$ images under the CC-BY license from Flickr, containing persons in self-contact. To obtain this data, we first crawl images with people by choosing a wide variety of tags (from daily activities to dance or sports) and run a person detector on the selection. Then, we pick images with persons in self-contact by manually classifying each person's bounding box in one of the 3 classes: "contact", "no contact" and "uncertain contact" (see fig.~\ref{fig:flickr_example}). To ensure pose variability among images with persons in self-contact we additionally run a 3d pose estimator \cite{zanfir_nips2018} and greedily select images that have a large 3d pose distance compared to the ones already selected. For the final pool of images we annotate the self-contact signature and image-space support of the signature, in a similar way to the HumanSC3D dataset. Statistics regarding the self-contact on the in-the-wild FlickrSC3D dataset can be visualized in fig.~\ref{fig:contact_statistics}.

\begin{table*}[!htbp]
\setlength{\tabcolsep}{0.8em} 
\begin{center}
\scalebox{0.9}{
\begin{tabular}{c|cc|cc|cc|cc}
& \multicolumn{2}{c}{\boldmath{$\text{IoU}_{75}$}} \vline &
\multicolumn{2}{c}{\boldmath{$\text{IoU}_{37}$}} \vline &
\multicolumn{2}{c}{\boldmath{$\text{IoU}_{17}$}} \vline &
\multicolumn{2}{c}{\boldmath{$\text{IoU}_{9}$}}\\
Dataset & Segm. &  Corresp. & Segm. &  Corresp. & Segm. &  Corresp. & Segm. &  Corresp. \\
\hline\hline
HumanSC3D & $0.469$ & $0.315$ & $0.560$ & $0.512$ & $0.703$ & $0.590$ & $0.787$ & $0.685$ \\
\hline
FlickrSC3D & $0.528$ & $0.422$ & $0.564$ & $0.475$ & $0.664$ & $0.579$ & $0.768$ & $0.692$ \\
\end{tabular}}
\end{center}
\caption{Annotator consistency as a function of the granularity of surface regions. }
\label{table:consistency}
\end{table*}
\noindent{\bf Annotator consistency.} For a small subset of images from both datasets, we ask two raters to annotate the self-contact signature. We check the annotator consistency at different levels of granularity of the body regions (from a fined-grained split into 75 regions to a coarser one, of only 9 body regions). We measure the intersection over union (IoU), first at a body region segmentation level, and then also by taking into account the set of correspondences between regions. Results are shown in table~\ref{table:consistency}. As in many other tasks, human annotation is not perfect, but it can be noticed that at smaller levels of granularity consistency increases and is of practical use (this certainly improves the quality of 3d reconstructions, as shown qualitatively and quantitatively in the following section).
\section{Experiments}
\noindent{\bf Self-Contact Image Support, Segmentation and Signature.} To assess the performance of our \ours{} network, we validate and test it on the FlickrSC3D dataset, which we split in the usual train ($80\%$), test ($10\%$) and validation ($10\%$) subsets. We train using the training set, validate our meta-parameters on the validation set and show the quantitative analysis on the test set. 

Since we are the first to propose explicitly learning the self-contact of the human body, there is no available method to compare against. The closest work in the literature is the ISP network \cite{fieraru2020cvpr} which predicts the contact signature between two humans in contact, which we adapt to learn a self-contact signature. We achieve it by removing one of the two computational pathways of ISP (the graph convolutional layers for one person and its respective specialization tasks for segmentation and signature learning) and then train and validate on the self-contact dataset FlickrSC3D. We train all methods on the finest granularity available ($N_{R}=75$), but also report results on multiple coarser granularities ($N_{R} = 37, 17 \text{ and } 9$), following the region splitting used in ISP for comparison. 

On both datasets, annotators have the freedom to choose whichever correspondences between facets of the human model and the image they prefer (as long as they are valid and the region-level self-contact segmentation is complete). This can lead to different multiple clicks in the image support of the same region. We set the ground truth image support of the respective region containing multiple coordinates as their average. Since the signature annotation is not necessarily complete (either because some regions are flagged \textit{masked}, when it is unclear in the image if they are involved in self-contact or not, or because just a subset of correspondences is annotated), we neither penalize them in training nor in evaluation. 

\begin{figure}
\begin{centering}
\includegraphics[height=0.14\textheight]{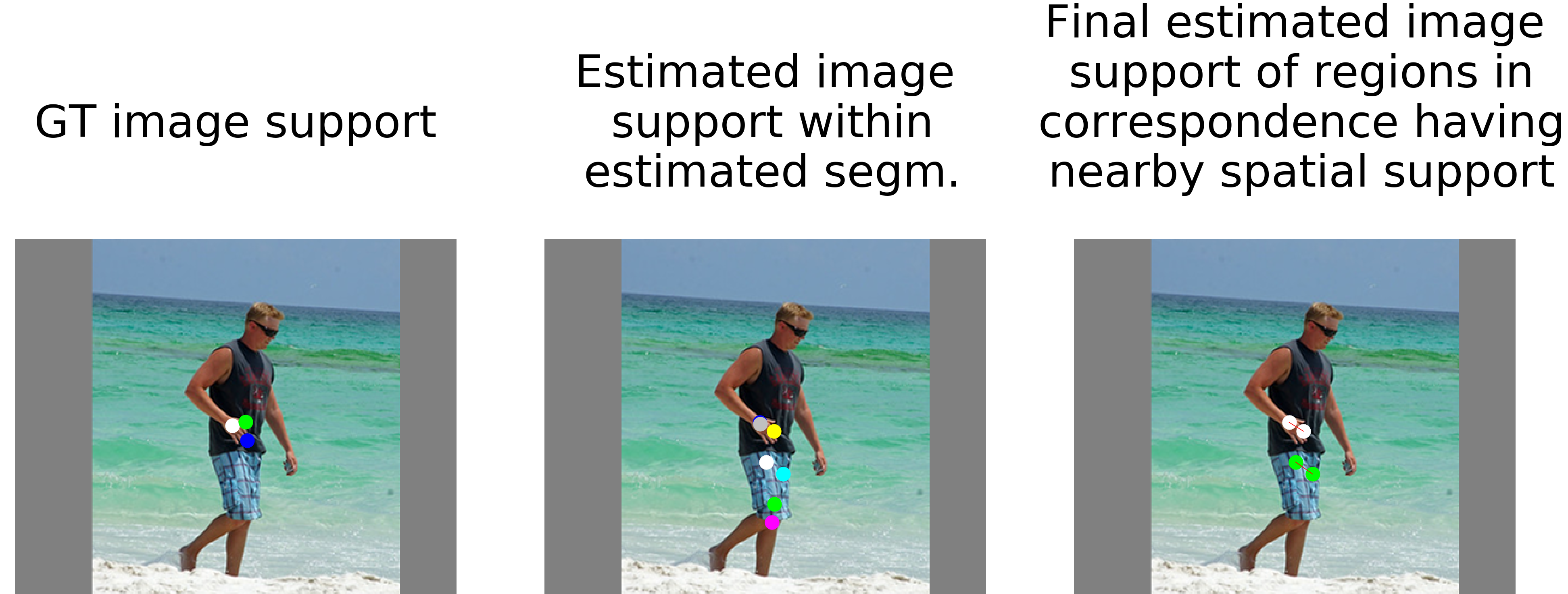} \\
\end{centering}
    \caption{\textbf{Left:} GT image self-contact support. \textbf{Center:} Contact image support for regions predicted to be in contact (within the segmentation prediction). \textbf{Right:} Estimated contact image support of regions found to be in correspondence and having nearby spatial support.}
\label{fig:image_support}
\end{figure}
\begin{table*}[!htbp]
\begin{center}
\scalebox{0.76}{
\begin{tabular}{|c|cc|cc|cc|cc|}\hline
& \multicolumn{2}{c}{\boldmath{$\text{IoU}_{75}$}} \vline &
\multicolumn{2}{c}{\boldmath{$\text{IoU}_{37}$}} \vline &
\multicolumn{2}{c}{\boldmath{$\text{IoU}_{17}$}} \vline &
\multicolumn{2}{c}{\boldmath{$\text{IoU}_{9}$}} \vline \\

\textbf{Method} 
& \textbf{Segm.} & \textbf{Signature} & \textbf{Segm.} & \textbf{Signature}  & \textbf{Segm.} & \textbf{Signature} & \textbf{Segm.} & \textbf{Signature}\\
\hline
\hline
\ours{} & $\mathbf{0.469}$ & $\mathbf{0.301}$ & $0.507$ & $\mathbf{0.339}$ & $0.591$ & $\mathbf{0.442}$ & $0.693$ & $\mathbf{0.550}$  \\ %
\hline
\ours{} w/o $\mathcal{L}_{sep}$ & $0.465$ & $0.289$ & $\mathbf{0.510}$ & $0.335$ & $0.603$ & $0.428$ & $0.692$ & $0.536$  \\ %
\ours{} w/o  $\mathcal{L}_{sep}$, w/o $\mathcal{L}_{K}$ & $0.460$ & $0.236$ & $0.502$ & $0.283$ & $\mathbf{0.605}$ & $0.395$ & $\mathbf{0.708}$ & $0.514$  \\ %
\hline
\makecell{\ours{} w/o imposing signature \\ consistency with image support} & $0.469$ & $0.244$ & $0.507$ & $0.288$ & $0.591$ & $0.395$ & $0.693$ & $0.501$  \\ %
\hline
\hline
ISP \cite{fieraru2020cvpr} (adapted for self-contact) & $0.462$ & $0.133$ & $0.503$ & $0.186$ & $0.583$ & $0.305$ & $0.688$ & $0.460$  \\ %
\hline
Human Consistency & $0.528$ & $0.422$ & $0.564$ & $0.475$ & $0.664$ & $0.579$ & $0.768$ & $0.692$ \\
\hline
\end{tabular}
}
\end{center}
\caption{Results of our self-contact segmentation and signature estimation on \textbf{FlickrSC3D}, evaluated for different region granularities on the human 3d surface (from 75, down to 9 regions). Human consistency is the same as in table~\ref{table:consistency}. We ablate the proposed losses and compare with the ISP baseline.}
\label{table:contact3d_quant_flickrsc3d}
\end{table*}
\begin{figure*}[!htbp]
\begin{center}
         \includegraphics[width=0.83\linewidth]{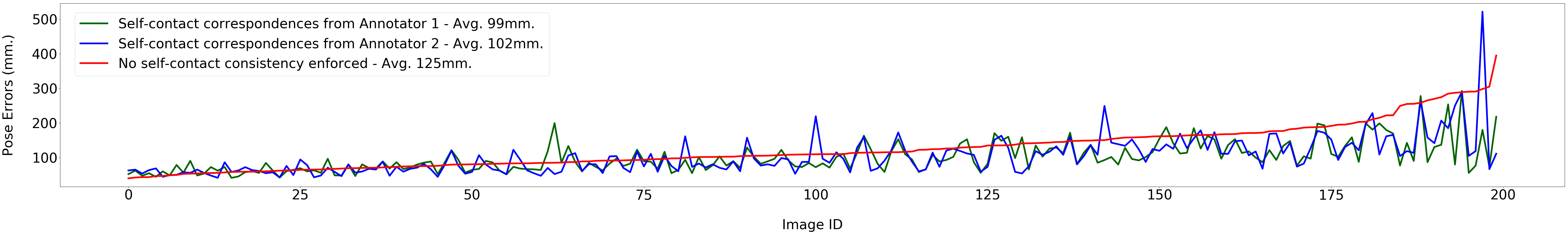}
\end{center}
\caption{Pose error for a subset of 200 images in HumanSC3D, each annotated by 2 human raters. We reconstruct the humans by enforcing contact consistency using the correspondences set by each annotator (\textit{green} and \textit{blue} line) and also without enforcing contact consistency (\textit{red} line, image IDs are ordered following this error). Enforcing contact consistency leads to smaller reconstruction error in $71.5\%$ cases (whether using correspondences from Ann. 1 or from Ann. 2), with $74\%$ agreement over the effect of enforcing contact consistency (either annotations improving or deteriorating the reconstruction simultaneously). The higher the initial pose error, the higher the improvement when enforcing self-contact consistency.}
\label{fig:ann_rec_err}
\end{figure*}

Table~\ref{table:contact3d_quant_flickrsc3d} shows quantitative results in terms of the intersection over union metric $\text{IoU}_{N_{R}}$, computed for different $N_{R}$. 
We show results for our method \ours{} and ablate different components. The strength of the proposed method is best seen at the finest level of granularity $N_{R}=75$, where it more than doubles the signature prediction performance, at $0.301$ vs. $0.133$ for ISP. The scenario where no additional supervision is used (for the self-contact image support) and the landmark discovery is unconstrained (\ours{} w/o  $\mathcal{L}_{sep}$ and $\mathcal{L}_{K}$) also outperforms ISP, showing that the effectiveness of our method does not only stem from the newly proposed self-contact image support annotations, but also from the more robust and specialized architecture. 
The effectiveness of the separability constraint $\mathcal{L}_{sep}$ is also shown for the self-contact signature task. 
Moreover, the experiment where the self-contact signature is not constrained to be consistent with the image support ($0.244$ $\text{IoU}_{75}$) shows the crucial, positive effect, of spotting and eliminating spurious correspondences by using the estimated image support. Fig.~\ref{fig:image_support} visualizes the image support estimates after applying different constraints.

\noindent{\bf Hand-Face Self-Contact}
To assess the possible application to analyzing disease transmission, we evaluate the hand-to-face self-contact detection. On our collected HumanSC3D, the hand-face correspondence is present in $34\%$ of images. On the problem of hand-to-face detection, \ours{} trained for general self-contact prediction obtains $46\%$ recall and $75\%$ precision. When retraining \ours{} with losses penalizing only the hand-to-face self-contact, the detection improves, to $53\%$ recall and $76\%$ precision. %
These are both major improvements from the $10\%$ recall and $66\%$ precision obtained by the adapted ISP baseline.
Still, the current results show that self-contact prediction is not a trivial problem, (see our Sup. Mat. video for failure cases), and can benefit from future research, potentially based on some of the methodology and data we provide. 

\noindent{\bf Self-Contact Signature for 3D Reconstruction}
To quantitatively assess the impact of self-contact consistency constraints in the quality of reconstructions, we test our method on the HumanSC3D dataset, where we report the MPJPE (mean per-joint position error) to evaluate the inferred pose, the translation error, the contact distance error (the minimum Euclidean distance between each pair of facets from two regions annotated to be in self-contact correspondence), and the per-vertex Euclidean distance error, measured against the (pseudo) ground truth meshes. Table~\ref{table:results_reconstruction} shows improvement across all the metrics when annotated self-contact signature is used to further constrain the reconstruction, showing that our annotations are valuable. Fig.~\ref{fig:ann_rec_err} plots the pose error for correspondences from two different annotators. While self-contact constraints do not always yield better reconstructions, on average, they do.

Fig.~\ref{fig:flickr-reconstructions} shows reconstruction results for images in FlickrSC3D, both with and without self-contact consistency constraints. By adding the penalty on annotated regions in correspondence, we recover accurate and visually plausible 3d reconstructions of challenging human poses. 
\begin{table*}[!htbp]
\begin{center}
\scalebox{0.68}{
\setlength{\tabcolsep}{4pt} 
\begin{tabular}{|c|cccc|cccc|cccc|cccc|}
\hline
 \textbf{Optim.} &
 \multicolumn{4}{|c|}{\textbf{W/o chair - standing}} & \multicolumn{4}{|c|}{\textbf{W/o chair - sitting}} & \multicolumn{4}{|c|}{\textbf{W/ chair}} &  \multicolumn{4}{|c|}{\textbf{Overall}} \\
  \textbf{loss} & 
  P & T & V & C &
  P & T & V & C &
  P & T & V & C &
  P & T & V & C \\

\hline
\hline
\textbf{ $L$ }  &  %
 $\mathbf{93.8}$ & $\mathbf{408.4}$ & $\mathbf{76.6}$  & $\mathbf{12.9}$ &
 $\mathbf{116.1}$ & $\mathbf{424.1}$ & $\mathbf{93.1}$ & $\mathbf{26.6}$ &
 $\mathbf{107.2}$ & $\mathbf{426.2}$ & $\mathbf{84.6}$ & $\mathbf{23.7}$ &
 $\mathbf{98.2}$ & $\mathbf{414.2}$ & $\mathbf{80.2}$ & $\mathbf{16.4}$ \\
 \hline
\textbf{$L$ w/o $L_G$ } & %
  $106.0$ & $419.3$ & $121.0$ & $210.2$ &
  $145.2$ & $436.0$ & $147.4$ & $182.7$ &
  $131.6$  & $431.9$ & $122.7$ & $189.3$ &
  $114.4$  & $423.4$ & $124.4$ & $203.4$ \\
 \hline
\end{tabular}
}
\end{center}
\caption{3D human \textbf{pose} (P), \textbf{translation} (T), \textbf{vertex} (V) estimation errors, as well as mean 3d \textbf{contact distance} (C), expressed in mm, for the HumanSC3D dataset. Using the full optimization function, with the geometric alignment term on annotated self-contact signatures, decreases the pose, translation and vertex estimation errors as well as the 3d distance between surfaces annotated as being in contact.}
\label{table:results_reconstruction}
\end{table*}
\begin{figure*}
[!htbp]
\def\hf{80pt}
\setlength{\tabcolsep}{0pt}
\begin{center}
\begin{tabular}{lcclcc}
         \includegraphics[height=\hf]{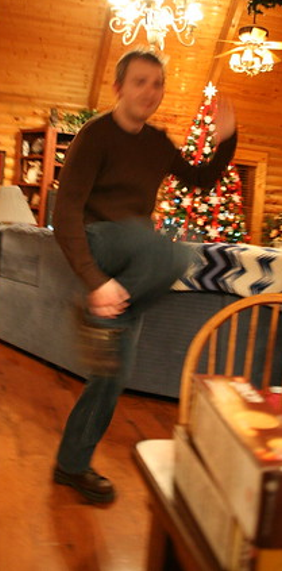} &
          \includegraphics[height=\hf]{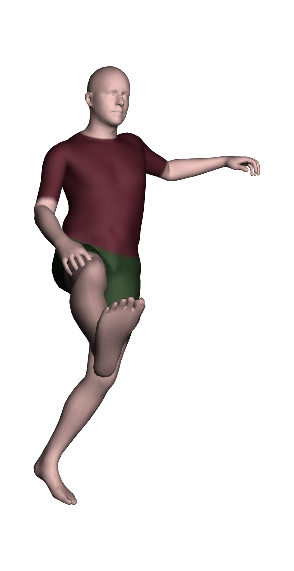} &
          \includegraphics[height=\hf]{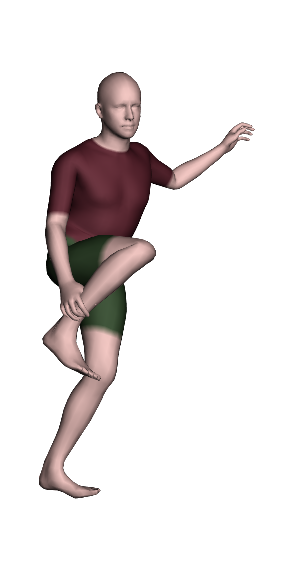} &
    
         \includegraphics[height=\hf]{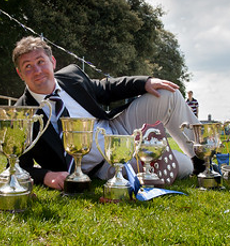} &
          \includegraphics[height=\hf]{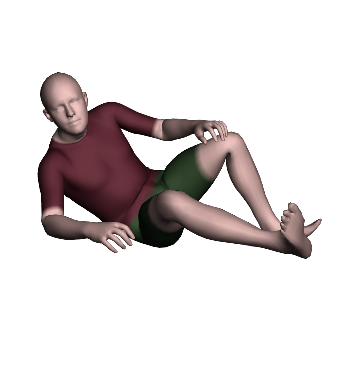} & 
            \includegraphics[height=\hf]{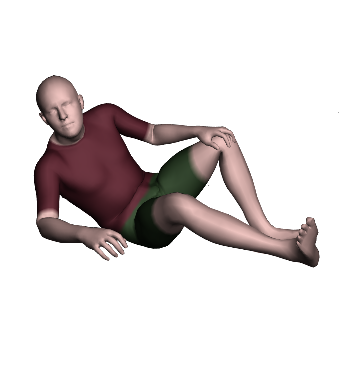} 
          \\
       \includegraphics[height=\hf]{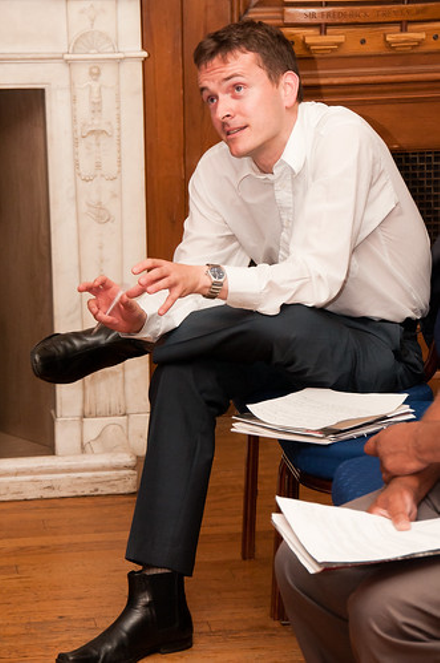} &
          \includegraphics[height=\hf]{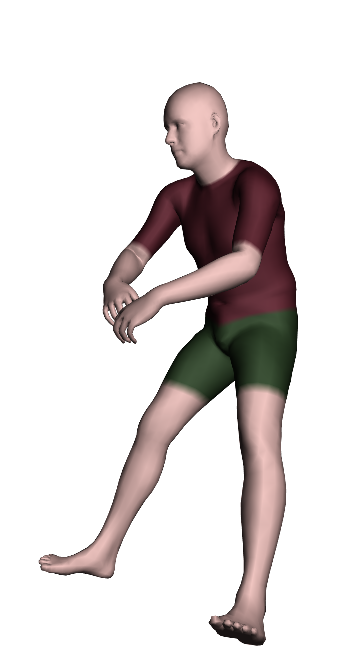} & 
            \includegraphics[height=\hf]{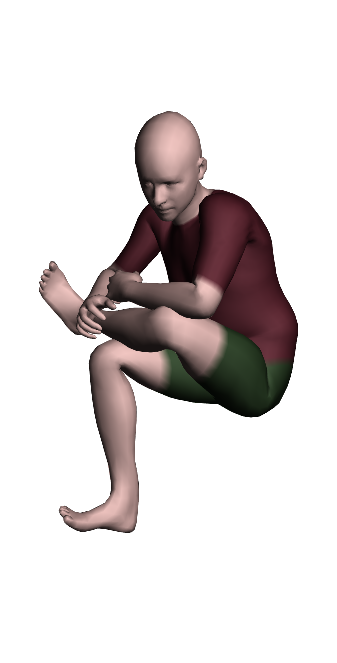}  & 
            
        \includegraphics[height=\hf]{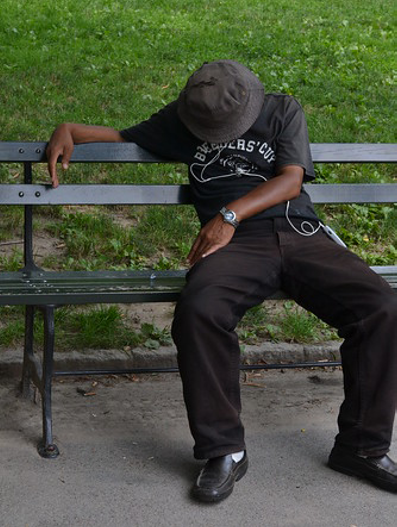} &
          \includegraphics[height=\hf]{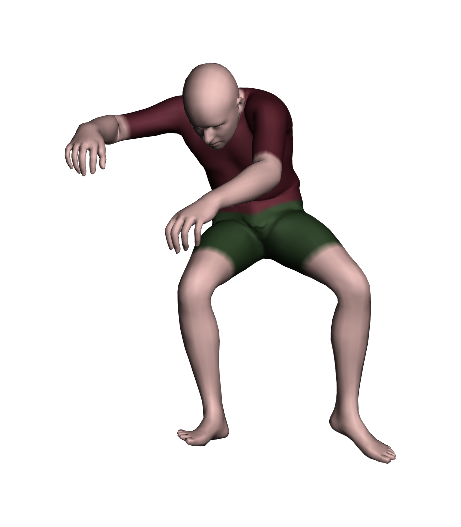} & 
            \includegraphics[height=\hf]{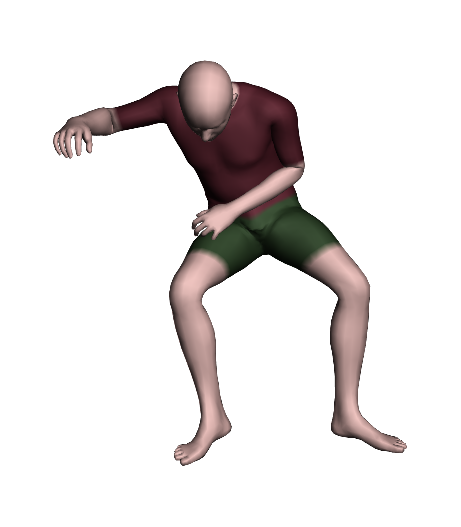} 
            \\
            
             \includegraphics[height=\hf]{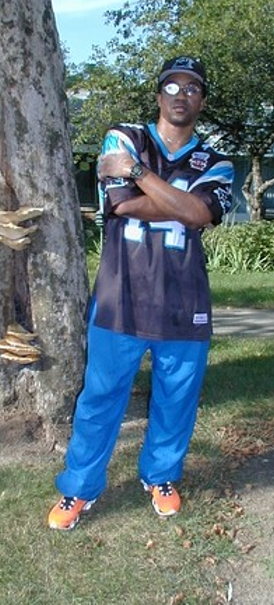} &
          \includegraphics[height=\hf]{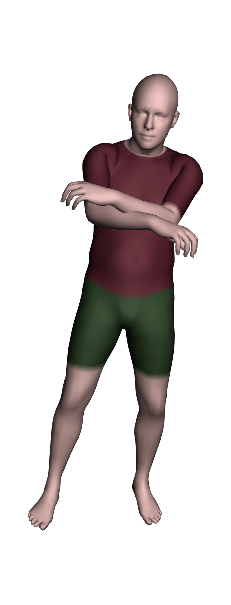} & 
            \includegraphics[height=\hf]{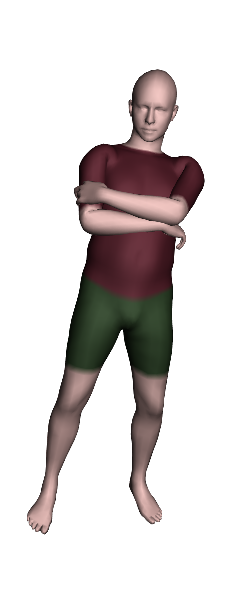} & 
        \includegraphics[height=\hf]{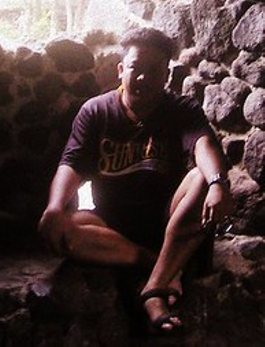} &
          \includegraphics[height=\hf]{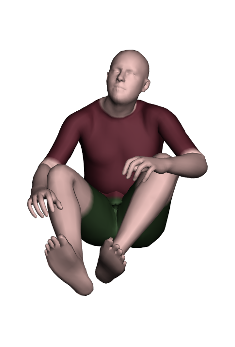} & 
            \includegraphics[height=\hf]{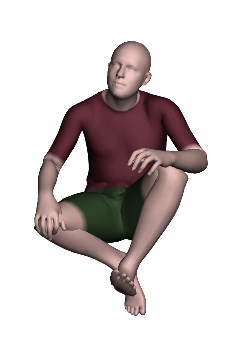} 
           
            \\
            \includegraphics[height=\hf]{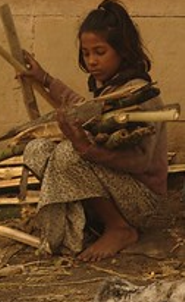} &
          \includegraphics[height=\hf]{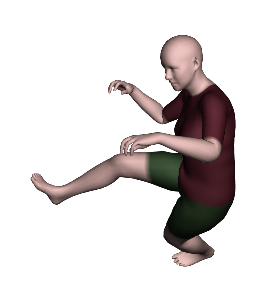} & 
            \includegraphics[height=\hf]{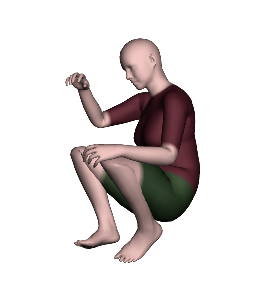} & 
            \includegraphics[height=\hf]{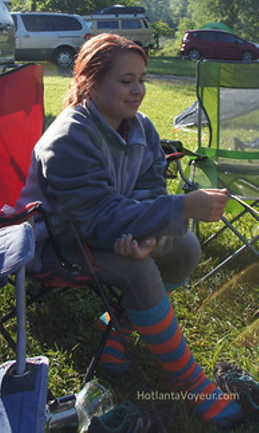} &
          \includegraphics[height=\hf]{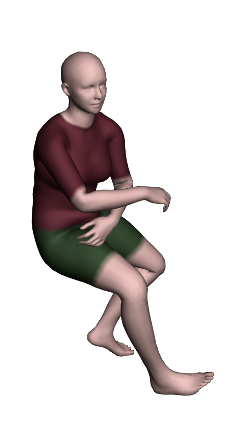} & 
            \includegraphics[height=\hf]{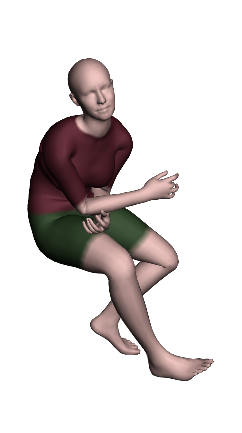}
            \\
                    \includegraphics[height=\hf]{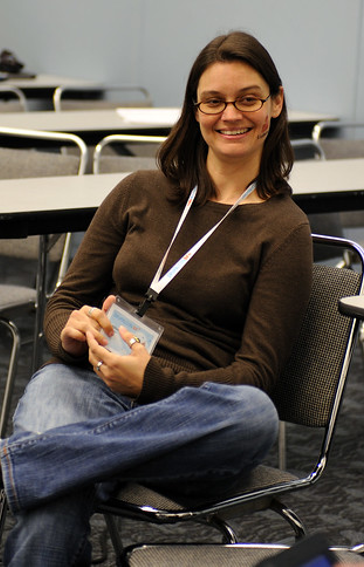} &
          \includegraphics[height=\hf]{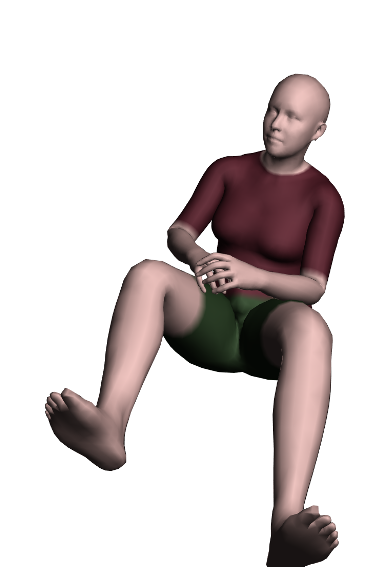} & 
            \includegraphics[height=\hf]{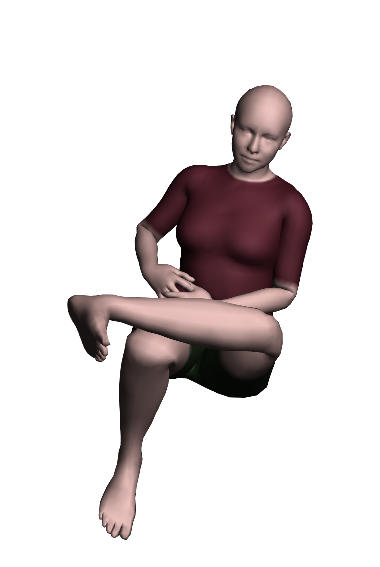}
            &
            \includegraphics[height=\hf]{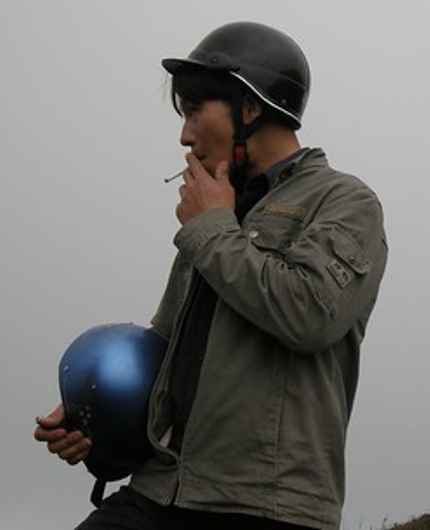} &
          \includegraphics[height=\hf]{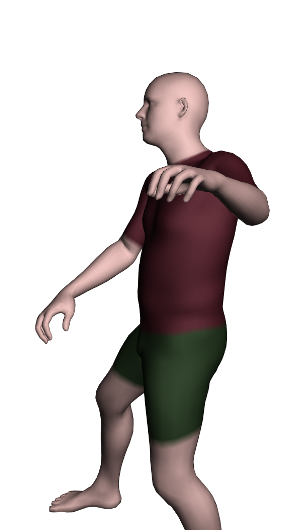} & 
            \includegraphics[height=\hf]{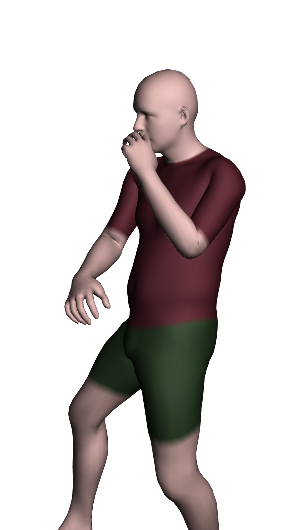}
            \\
          \includegraphics[height=\hf]{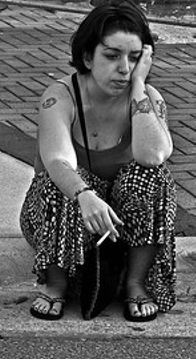} &
          \includegraphics[height=\hf]{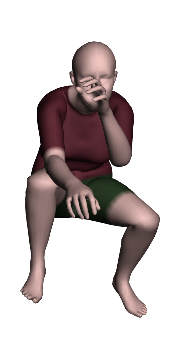} & 
            \includegraphics[height=\hf]{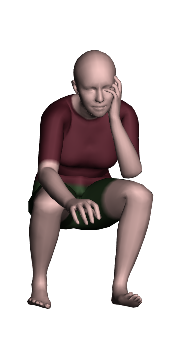} & 
         \includegraphics[height=\hf]{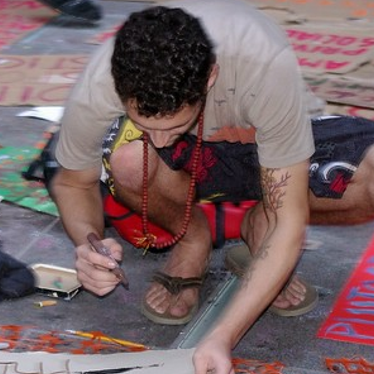} &
          \includegraphics[height=\hf]{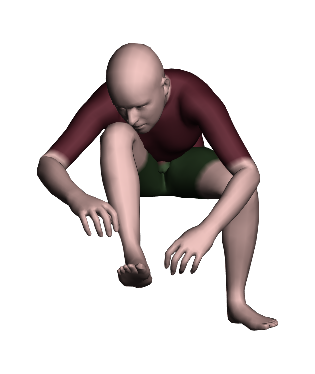} & 
            \includegraphics[height=\hf]{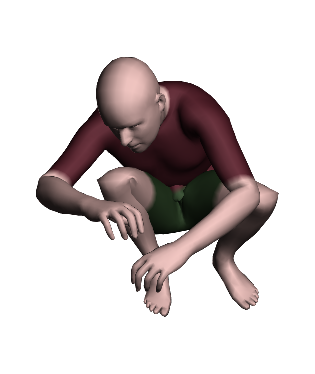}

         \end{tabular}

\end{center}
\caption{3D pose and shape reconstructions using our annotated self-contact data. \textbf{Left}: Original image. \textbf{Center}: Reconstruction without considering the self-contact and the associated loss. \textbf{Right}: Reconstruction that uses the self-contact annotations and the corresponding loss. }
\label{fig:flickr-reconstructions}
\end{figure*}

\section{Conclusions}
We have presented the task of human self-contact estimation and the design of the \ours{} methodology to detect body surface regions in self-contact, the correspondences between them, and their spatial support. By integrating this methodology with 3d explicit self-contact losses, we have shown that 3d visual reconstructions of human self-touch events are possible with superior quantitative and perceptual results over non-contact baselines. 
The models we built had their component effectiveness evaluated based on a large dataset collected in the wild, containing $25,297$ image-surface-surface correspondence annotations, as well as a motion capture dataset containing $5,058$ contact events and $1,246,487$ ground truth 3d poses. This represents a considerable amount of logistic, collection and annotation work, involving human subjects, and will be made available to the research community.\footnote{ \texttt{http://vision.imar.ro/sc3d}} Finally, we have demonstrated an application to detecting face-touch and showed how self-contact signatures can enable more expressive 3d reconstruction, thus opening path for subtle 3d behavioral reasoning in the future.

\noindent{\textbf{Acknowledgments:}} This work was supported in part by the ERC Consolidator grant SEED, CNCS-UEFISCDI (PN-III-P4-ID-PCCF2016-0180) and SSF.

\section{Supplementary Material}
In this supplementary material, we include implementation details for our proposed self-contact prediction method \ours{} and methodological details on the 3d reconstruction method. In addition, we show evaluation results for the self-contact image support prediction. For a demo of the annotation process, as well as qualitative results of \ours{} and the 3d reconstruction method (on both FlickrSC3D and HumanSC3D), please check our Sup. Mat. video on our project page. 

\subsection{Self-Contact Prediction - Implementation Details}
All hyper-parameters of the \ours{} network and its associated losses were validated on the validation subset of FlickrSC3D.
$\Theta_K$ consists of a convolutional layer and a batch normalization layer that reduces the number of channels from $512$ to $N_{R}=75$. Softargmax is defined as ${Softargmax}(x) = (\sum_{i=1}^{W}\sum_{j=1}^{H} \frac{i}{W}\sigma(x)_{i,j}, \sum_{i=1}^{W}\sum_{j=1}^{H} \frac{j}{H}\sigma(x)_{i,j})$, where $\sigma(x)$ is the softmax normalization function and $W,H$ are the width and height of the input $x$. $\Theta_{agg}$ consists of a linear layer that reduces the dimension of each feature from $1024$ to size $60$, followed by a fully connected layer, aggregating the features and reducing their dimension $N_{R} \times 60 \rightarrow N_{R} \times 20$. 
$\Theta_S$ and $\Theta_C$ are both fully connected layers outputting the last features $F$ of size $N_{R} \times 10$. In $\mathcal{L}_{sep}$, we find $\sigma^2_{sep} = 0.025$ to be optimal, given $0 \leq \widetilde{x_{r}}, \widetilde{y_{r}} \leq 1$. The total loss used to train the model is a weighted sum of each loss: $\mathcal{L} = w_{sep} \mathcal{L}_{sep} + w_K \mathcal{L}_{K} + w_S\mathcal{L}_{S} + w_K\mathcal{L}_{C}$, with $w_{sep} = w_{K} = 5.0$ and $w_{S} = w_{C} = 1.0$. 
  
We preprocess the input RGB image by rescaling it such that the height of the person is fixed ($220$px), where \cite{cao2018openpose} is used as a person detector. We also follow the data augmentation procedure proposed in \cite{cao2018openpose}: random scaling by a $50-110\%$ factor, cropping randomly within $40$px around the center of the person and flipping randomly around the vertical axes. The input crop is fixed to $368$px $\times$ $368$px.

To train the network, we use stochastic gradient descent with a mini-batch of 15 images. We use an initial learning rate of $10^{-3}$, with a momentum of $0.9$. The learning rate decreases when the loss on the validation set plateaus. Training is performed for 80 epochs and the model selection is done to minimize the validation loss.

We validate the thresholding of $S$ for maximum intersection over union with the ground-truth self-contact segmentation. Similarly, we validate two thresholds, one for removing correspondences in $C$ that have small probability, and another one to further remove those whose corresponding regions are far from each other. 

The implementation is done in the open-source PyTorch \cite{paszke2017automatic} framework.

\subsection{Self-Contact Signature for 3D Reconstruction}
We describe here how self-contact signature $C^{R}$ can be used to constrain 3d human reconstruction. We use the optimization framework of \cite{Zanfir_2018_CVPR} augmented with the contact consistency loss $L_{G}$ proposed in \cite{fieraru2020cvpr}, changed to enforce self-contact consistency for one person instead of imposing interaction contact between two people. 

Our goal is to find the 3d pose and shape of a person in an image. We 
initialize the optimization with the predicted mesh parameters of \cite{zanfir_nips2018}. We then iterate to minimize the cost function $L$ by propagating gradients from the loss function through the 3d body model. The cost function is: 
\begin{equation}
\label{eq:full_loss}
  L = L_{S} + L_{psr} + L_{col} + L_{G}
\end{equation}

$L_{S}$ is the projection error with respect to estimated semantic body part labeling and 2d body pose, $L_{psr}$ is a pose and shape regularization cost, and $L_{col}$ is a self-collision penalty term. 
The self-contact consistency loss $L_{G} = L_{D} + L_{N}$ contains two terms: $L_{D}$, minimizing the distance between pairs of regions in self-contact and $L_{N}$, aligning their orientation.

$L_{D}$ is defined as the sum of distances between regions in self-contact ($r_1, r_2$ for which $C_{r_1, r_2}^{R} = 1$):
\begin{align}
    L_{D} = \sum_{\substack{(r_1, r_2) \text{ s.t.} \\ C_{r_1, r_2}^{R} = 1}} \Phi_D(r_1, r_2)
\end{align}
where the distance between two regions on the same body $\Phi_D(r_1, r_2)$ is the sum of Euclidean distances between each facet in one region and its nearest-neighbour facet in the opposite region:
\begin{align}
\label{eq:PhiD}
    \Phi_D(r_1, r_2) = &\sum_{f_1\in \psi_D(r_1)}\min_{f_2 \in \psi_D(r_2)} \phi_D(f_1,f_2) + \\
    \nonumber
    &\sum_{f_2\in \psi_D(r_2)}\min_{f_1 \in \psi_D(r_1)} \phi_D(f_1,f_2)
\end{align}
The Euclidean distance $\phi_D(f_1, f_2)$ is computed between the centers of two facets $f_1, f_2$. $\psi_D(r)$ can select all facets of region $r$, a subset of them, or the facet in the center of the region. 

$L_{N}$ is defined as the sum of facet orientations between the nearest-neighbour matches ($\psi_N$) found in eq.~\ref{eq:PhiD} .
\begin{equation}
    L_{N} = \sum_{\substack{(r_1, r_2) \text{ s.t.} \\ C_{r_1, r_2}^{R} = 1}} \Bigg(\sum_{(f_1, f_2) \in \psi_N(r_1, r_2)} {N(f_1)} \bullet {N(f_2)}\Bigg)
\end{equation}
where $N(f)$ is the unit-length surface normal of facet $f$. Note that $L_{N}$ is minimized when the angle between corresponding normals is $\pi$.

In our main paper, we showed reconstruction results for both cases: when $L_{G}$ is included in $L$, and when it is not. Their comparison shows the benefits, qualitative and quantitative, of enforcing self-contact consistency using the $L_{G}$ term.

\subsection{Self-Contact Image Support - Evaluation}
Here we evaluate the estimation of self-contact image support. For the regions in the ground-truth self-contact segmentation, we compute the average Euclidean distance between the predicted image support and the annotated support. Results are reported in Table~\ref{table:spatial_support_err} for our full method (\ours{}), our method without the separation loss (\ours{} w/o $\mathcal{L}_{sep}$) and the version without self-contact image support supervision (\ours{} w/o  $\mathcal{L}_{sep}$, w/o $\mathcal{L}_{K}$).
\begin{table}[!htbp]
\setlength{\tabcolsep}{0.8em} 
\begin{center}
\scalebox{0.85}{
\begin{tabular}{c|cc}
& \multicolumn{2}{c}{Error (in pixels)} \\
Method & FlickrSC3D & HumanSC3D\\
\hline\hline
\ours{} & $13.6$ & $10.3$\\ %
\hline 
 \ours{} w/o $\mathcal{L}_{sep}$ & $12.8$ & $9.2$\\ %
\hline
\ours{} w/o  $\mathcal{L}_{sep}$, w/o $\mathcal{L}_{K}$ & $70.6$ & $61.0$\\ %

\end{tabular}}
\end{center}
\caption{Self-contact image support error on the test set of FlickrSC3D and HumanSC3D. We report the Euclidean distance (in pixels), where images have size $368$px $\times$ $368$px. Note that supervising self-contact image support (\ours{} and \ours{} w/o $\mathcal{L}_{sep}$) leads to tiny errors, showing that the support was in fact learned, and can be used for downward tasks, such as self-contact signature estimation.}
\label{table:spatial_support_err}
\end{table}

\section*{Broader Impact and Ethics}
While research is still experimental, in the long run our models can be integrated into systems performing large-scale psycho-social studies of human behavior. Such models can also be used with personal assistants that can operate under high privacy standards and can be accountable to humans. Assistants can rely on self-contact signatures to reason about a person's internal state including emotional response, and could provide feedback to that person over a period of time, for increase awareness or for positively changing habits. Our work can also be potentially relevant to detect and correct the unconscious behaviour of touching one's mouth or face with the hand \cite{kwok2015face}, in order to avoid infections with various pathogens, if proper hygiene is not maintained. In this regard, the models can be potentially used in hospitals in order to monitor hygiene of both patients and medical personnel. The work can also be potentially applied in the monitoring and treatment of people with a history of self-harm  \cite{hawton2015self}. During data collection, we aimed to reduce bias, by having a diverse and representative collection of humans in self-contact, within our limited subject and annotation budget.

\bibliography{human.bib}
\end{document}